\documentclass[10pt,twocolumn,letterpaper]{article}
\usepackage[pagenumbers]{cvpr}
\usepackage{listings}
\usepackage{xcolor}

\lstdefinestyle{vlmprompt}{
    basicstyle=\ttfamily\footnotesize,
    breaklines=true,
    breakatwhitespace=false,
    columns=fullflexible,
    keepspaces=true,
    showstringspaces=false,
    frame=single,
    backgroundcolor=\color{gray!5}
}
\usepackage[T1]{fontenc}
\usepackage[utf8]{inputenc}
\usepackage{microtype}
\usepackage{array,tabularx,multirow,makecell}
\usepackage{mathtools}
\usepackage{adjustbox}
\usepackage{placeins}

\graphicspath{{figures/main/}{figures/supplementary/}}

\definecolor{cvprblue}{rgb}{0.21,0.49,0.74}
\usepackage[pagebackref,breaklinks,colorlinks,allcolors=cvprblue]{hyperref}
\usepackage{comment}
\def\confName{CVPR}
\def\confYear{2027}
\title{Bilateral Evidence Selection and Reasoning\\
for Mammography Vision--Language Models}
\author{
  Zhengbo Zhou$^{1}$ \quad
  Dooman Arefan$^{2}$ \quad
  Shandong Wu$^{1,2,3}$\\
  $^{1}$Intelligent Systems Program \quad
  $^{2}$Department of Radiology\\
  $^{3}$Department of Biomedical Informatics and Department of Bioengineering\\
  University of Pittsburgh, Pittsburgh, PA, USA
}

\begin{document}
\maketitle

\begin{abstract}
Medical vision--language models (VLMs) often need to identify small, subtle lesions within high-resolution images containing extensive normal anatomy and redundant visual information. Visual-token compression improves efficiency, but preserving salient regions alone can discard the context needed for diagnostic interpretation. This context can extend across images: in bilateral mammography, the opposite breast provides a patient-specific anatomical reference. Yet retaining such evidence does not ensure that the model uses it effectively. We propose Bilateral Evidence Selection and Reasoning (BESR), which connects evidence preservation with comparative assessment through a bilateral-guided token selector and a bilateral reward. Under a shared token budget, the selector jointly prioritizes lesion appearance and its local and bilateral context, while the reward encourages assessments that benefit from the preserved comparison. Experiments on a bilateral dataset constructed from CBIS-DDSM and on VinDr-Mammo demonstrate improved evidence retention and diagnostic performance under constrained visual-token budgets. Inference benchmarking further shows reduced prefill latency.
\end{abstract}

\section{Introduction}
\label{sec:introduction}

Medical vision--language models (VLMs) often need to identify small, subtle lesions within high-resolution images containing extensive normal anatomy and redundant visual information~\cite{li2026medreasonr1,park2026medicad}. Although the evidence relevant to a diagnostic question may be concentrated in a few regions, the model processes a long visual sequence representing the entire image. This increases inference cost and requires the language model to identify subtle findings among many less relevant regions. Efficient medical VLMs therefore need selective visual representations that preserve the evidence required for lesion characterization and assessment.

Visual-token compression offers a way to reduce this burden. By selecting, pruning, or aggregating visual tokens, existing methods shorten the sequence processed by the language model and reduce computation while preserving much of the task-relevant information~\cite{chen2024fastv,yang2025visionzip}. Selection criteria include attention, text-guided importance, feature diversity, and multimodal coverage~\cite{zhang2025sparsevlm,alvar2025divprune,dong2026mmtok}. Other approaches incorporate contextual regions, spatial structure, or clinical concepts~\cite{guo2025crop,huang2026nuwa,liu2026breastgpt}. However, retaining a compact set of informative tokens does not by itself resolve the evidence requirements of medical interpretation. A salient lesion region may survive compression while less salient tissue needed to characterize its boundary or provide a reference is discarded.
\begin{figure}[t]
\centering
\includegraphics[width=\columnwidth]{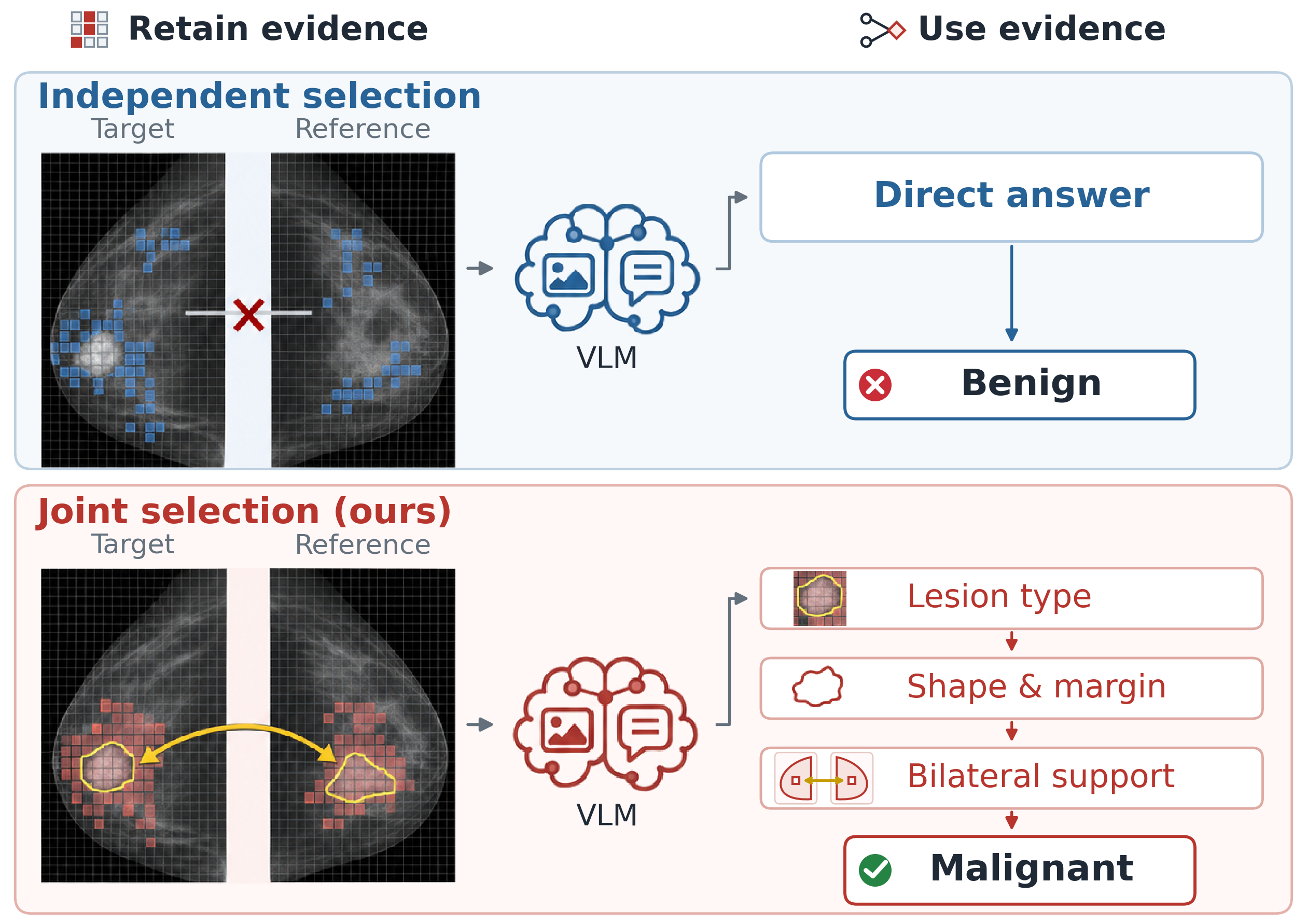}
\caption{\textbf{Two motivations for our framework.}
(1) Use bilateral asymmetry to guide token selection.
(2) Interpret lesion morphology and contralateral appearance to support
benign--malignant classification.}
\label{fig:motivation}
\end{figure}
The importance of such reference information is reflected in medical VLMs that incorporate related images. MAIRA-2 uses current and prior radiographs together with multiple views for report generation~\cite{bannur2024maira2}; MEDIC-AD explicitly represents temporal differences between examinations~\cite{park2026medicad}; and MammoRG integrates four-view mammograms for reporting~\cite{zhu2026mammorg}. These approaches establish the value of incorporating information across images. They also motivate considering which relationships remain available when those images are compressed: a region with limited relevance in isolation can provide useful evidence through comparison with another region.

Bilateral mammography makes this requirement particularly clear. Interpreting a lesion involves its local appearance, boundary, and surrounding tissue, while the opposite breast supplies patient-specific anatomical context for assessing symmetry and focal differences~\cite{liu2019cbn,liu2022agn,wang2023disasymnet}. The value of a contralateral reference therefore depends on the finding it helps interpret, rather than solely on its own visual salience. Selecting tokens independently across the two breasts can retain a suspicious region while discarding its matched reference and concentrating only on the lesion can also remove surrounding context. As illustrated in Fig.~1, a useful compact representation should preserve these complementary regions together under the available budget. 

We propose \textbf{Bilateral Evidence Selection and Reasoning (BESR)}, a framework that couples a \emph{bilateral-guided token selector} with a \emph{bilateral reward} to connect evidence preservation with lesion characterization and comparative assessment. The selector jointly allocates a shared visual-token budget across the two breasts, prioritizing lesion appearance, spatial context, and contralateral support. A target--reference pair contributes comparative value only when both regions are retained, with uncertain correspondences downweighted. This makes the relationships needed for comparison an explicit consideration during compression.

Preserving evidence, however, does not ensure that the VLM interprets it correctly: a correct assessment can coexist with an inaccurate description of the lesion. Recent medical RL methods demonstrate the value of correctness-based post-training~\cite{lai2026medr1,li2026medreasonr1,huang2026medvlthinker}. However, a correct assessment does not necessarily imply that the accompanying lesion description is accurate. BESR therefore trains the VLM to predict lesion type and clinically defined attributes, such as shape and margin, alongside its assessment and a concise explanation. Explicit correctness rewards for the descriptor and assessment labels make lesion characterization a learning objective. The bilateral reward complements this supervision by evaluating whether the preserved reference supports the assessment. Conditioned on the same model-generated attributes, it favors correct assessments that receive stronger support from intact than spatially shuffled reference features. We combine these rewards within group relative policy optimization (GRPO)~\cite{shao2024deepseekmath}, keeping the visual evidence module frozen so that post-training focuses on interpreting the retained evidence.

We evaluate BESR on a bilateral dataset constructed from CBIS-DDSM and on VinDr-Mammo, examining evidence retention, diagnostic performance, and inference efficiency across visual-token budgets. Our contributions are threefold:
\begin{itemize}
    \item We propose a \textbf{bilateral-guided token selector} that jointly prioritizes lesion appearance, spatial context, and contralateral support under a shared budget, explicitly valuing the retention of complementary evidence for comparison.

    \item We introduce a \textbf{bilateral reward} that evaluates the additional support provided by a matched reference for a correct assessment, connecting policy optimization to the comparative evidence retained by the selector.

    \item We demonstrate \textbf{diagnostic gains} on bilateral CBIS-DDSM and VinDr-Mammo, together with \textbf{improved inference efficiency}. At $K=256$, BESR improves balanced accuracy and F1 on CBIS-DDSM by 8.09 and 10.97 percentage points over the strongest evaluated 256-token medical reasoning baselines for the respective metrics. At the same budget, our inference benchmark shows 87.1\% fewer visual tokens and a $4.52\times$ prefill speedup relative to full-token inference.
\end{itemize}

\section{Related Work}
\label{sec:related-work}

\paragraph{Efficient visual representations.}
Visual-token compression reduces VLM computation through token pruning and aggregation. FastV uses early decoder attention to identify tokens for pruning in subsequent layers, while SparseVLM combines text-guided importance with redundancy reduction~\cite{chen2024fastv,zhang2025sparsevlm}. DivPrune prioritizes feature diversity, MMTok jointly optimizes visual and textual coverage, and VisionZip combines dominant-token selection with contextual aggregation~\cite{alvar2025divprune,dong2026mmtok,yang2025visionzip}. Beyond individual token importance, CROP localizes query-relevant contextual regions, and N\"uwa preserves spatial structure for visual grounding~\cite{guo2025crop,huang2026nuwa}. These methods motivate compact representations that preserve task-relevant information. BESR focuses on the joint value of finding-specific evidence: its selector jointly prioritizes local appearance, surrounding context, and contralateral support under a shared budget, providing the comparisons that subsequent policy optimization is trained to use.

\paragraph{Bilateral comparison and medical grounding.}
Bilateral and multi-view mammography models exploit anatomical relationships and cross-breast comparison for lesion analysis. CBN and AGN model complementary information across breasts and views~\cite{liu2019cbn,liu2022agn}, while DisAsymNet disentangles asymmetrical abnormalities from symmetric anatomy~\cite{wang2023disasymnet}. MammoRG extends multi-view modeling to mammography reporting through clinical knowledge integration~\cite{zhu2026mammorg}. More broadly, MAIRA-2, MIMO, and VividMed associate medical language with localized image evidence through grounded reporting, visual referring, and spatial outputs~\cite{bannur2024maira2,chen2025mimo,luo2025vividmed}. MEDIC-AD incorporates anomaly-aware and temporal-difference representations to support lesion analysis and comparison~\cite{park2026medicad}. These studies establish the importance of spatial and relational evidence. BESR investigates which finding--reference relationships remain available after compression and uses the retained visual support for subsequent description and assessment.
\begin{figure*}[t]
  \centering
  \includegraphics[width=\textwidth]{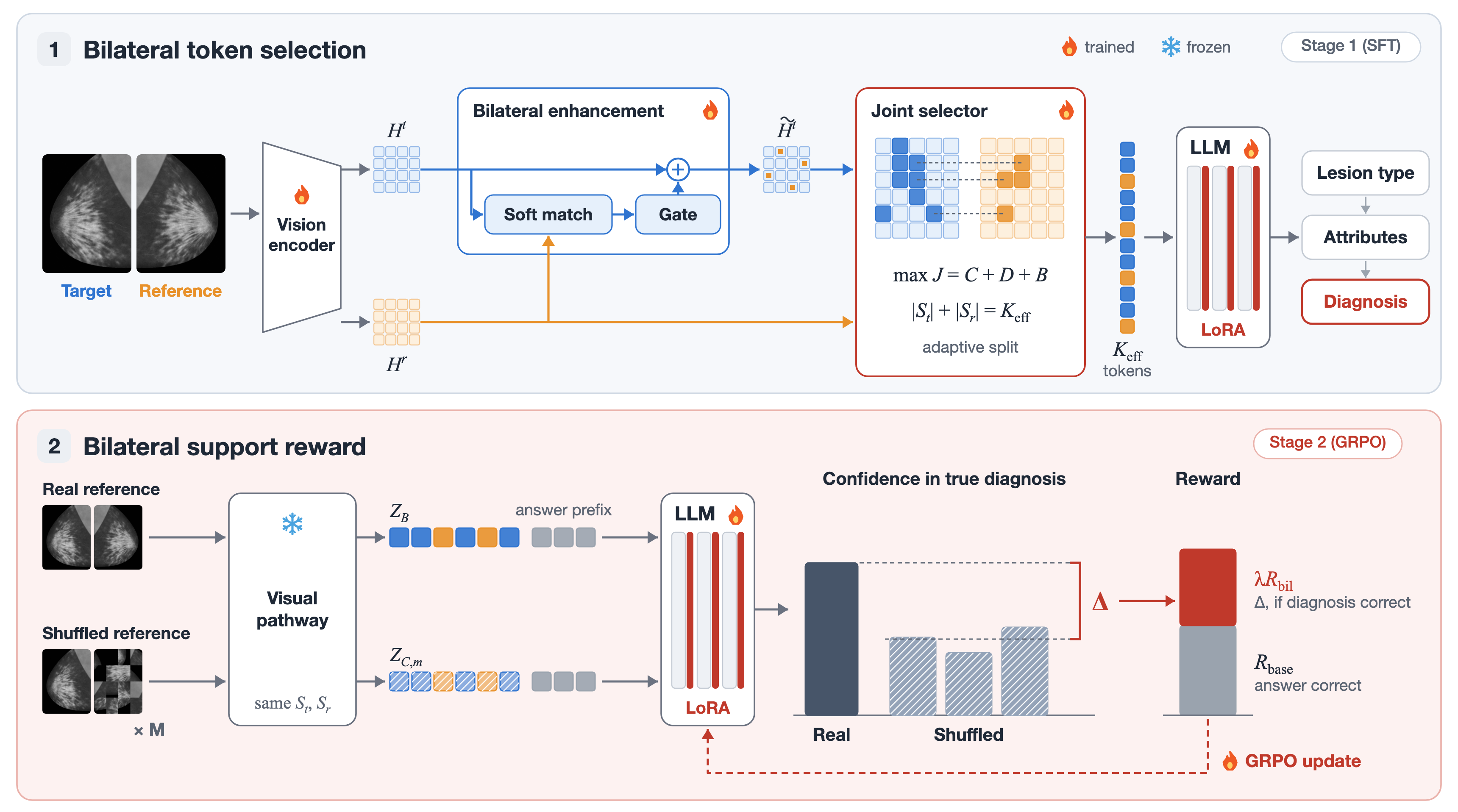}
  \caption{\textbf{Bilateral Evidence Selection and Reasoning (BESR).}
(1)~Bilateral enhancement pools matched reference context into target features,
and the joint selector allocates $K_{\text{eff}}$ tokens across both breasts by
maximizing coverage $C$, local relevance $D$, and pair utility $B$. The decoder
predicts lesion type, attributes, and diagnosis.
(2)~During GRPO, with the visual pathway frozen, a correct response receives a
bounded bonus when the intact reference supports its diagnosis more than
shuffled controls. }
  \label{fig:framework}
\end{figure*}
\paragraph{Clinical concepts and reasoning post-training.}
Structured concepts provide an interface between visual findings and predictions. Concept Bottleneck Models predict explicit intermediate concepts~\cite{koh2020cbm}; Deep BI-RADS combines supplied clinical descriptors with mammograms~\cite{benartzi2025deepbirads}; and MedCBR connects learned concepts to guideline-based explanations~\cite{Harmanani_2026_CVPR}. Complementary work improves the reasoning process itself. Visual CoT explores region-focused reasoning~\cite{shao2024visualcot}.
Med-R1 uses GRPO for generalizable medical reasoning~\cite{lai2026medr1},
MedReason-R1 combines local zooming with RL for CT
diagnosis~\cite{li2026medreasonr1}, and MedVLThinker studies
post-training with verifiable answer rewards~\cite{huang2026medvlthinker}.
ViTAR combines iterative visual grounding with RL~\cite{chen2025think}.
Beyond correctness and output-format rewards, BESR rewards stronger
support for correct assessments from intact versus shuffled
contralateral evidence.

\section{Method}
\label{sec:method}

\subsection{Overview}
Given a target mammogram $I^t$ and a contralateral reference $I^r$
from the same examination and projection, BESR selects a compact set
of visual tokens and generates lesion descriptors, a diagnostic
assessment, and a concise explanation. As shown in Fig.~2, our
\emph{bilateral-guided token selector} allocates one budget across
both breasts, retaining informative target regions
together with useful reference regions. Supervised learning trains
the selector and structured prediction. We then freeze the visual
pathway and optimize decoder LoRA parameters with rewards for
descriptor and assessment correctness, together with a
\emph{bilateral diagnostic-support reward}. This additional reward
favors correct assessments that receive stronger support from intact
than disrupted reference context, given the model's own descriptors.

\subsection{Bilateral-Guided Token Selection}
\label{sec:encoding}

\paragraph{Reference-conditioned target features.}
A shared Qwen2.5-VL vision encoder and visual merger~\cite{bai2025qwen25vl}
produce target tokens $H^t=E_\phi(I^t)$ and reference tokens
$H^r=E_\phi(I^r)$, with lengths $N_t$ and $N_r$.
After aligning image orientations, each target token searches a local
reference neighborhood $\mathcal N(i)$. Soft weights $a_{ij}$ combine
feature similarity and spatial proximity, accommodating approximate
correspondence between the breasts. We aggregate reference context
and update the target features:
\begin{equation}
\begin{aligned}
\bar h_i^r &= \sum_{j\in\mathcal N(i)}a_{ij}h_j^r,\\
\tilde h_i^t &= h_i^t+g_i\,\psi(h_i^t,\bar h_i^r).
\end{aligned}
\label{eq:bilateral-enhancement}
\end{equation}
The learned update $\psi$ uses target appearance, reference context,
and their difference. The gate $g_i$ controls the update using these
features and the concentration of the soft match. The residual
connection provides a direct path for target features. Matching and
gating details are given in the supplement.

\paragraph{Joint selection objective.}
The selector preserves relevant target regions and allocates
reference tokens that support their comparison.
Let $\widetilde H^t=\{\tilde h_i^t\}_{i=1}^{N_t}$ and
$H^r=\{h_j^r\}_{j=1}^{N_r}$ denote the enhanced target and
reference features, with $N_t$ and $N_r$ tokens, respectively.
A learned relevance head $f_{\mathrm{rel}}$ predicts
\begin{equation}
z_i=f_{\mathrm{rel}}(\tilde h_i^t),
\qquad
p_i=\frac{\exp(z_i)}
{\sum_{u=1}^{N_t}\exp(z_u)},
\label{eq:target-relevance}
\end{equation}
where $z_i$ is the unnormalized score and $p_i$ is the
normalized relevance of target token $i$.
Lesion annotations supervise these scores during training.
At inference, relevance is predicted from image features. We select target indices $S_t$ and reference indices $S_r$
under a shared budget:
\begin{equation}
\begin{aligned}
J(S_t,S_r)
&=\underbrace{C(S_t)+D(S_t)}_{\text{target evidence}}
+\underbrace{B(S_t,S_r)}_{\text{bilateral support}},\\
|S_t|+|S_r|&=K_{\mathrm{eff}},
\qquad
K_{\mathrm{eff}}=\min(K,N_t+N_r).
\end{aligned}
\label{eq:joint-selection}
\end{equation}
Here, $K$ is the requested token budget,
$K_{\mathrm{eff}}$ is the number of tokens retained,
and $|\cdot|$ denotes set cardinality.

\paragraph{Target coverage and local relevance.}
Coverage $C$ measures how much of the predicted target
relevance map is represented.
Let $\mathcal M(i)$ denote the immediate neighborhood of
target token $i$ on the spatial grid.
A selected location receives full credit, while an
unselected neighboring location receives partial credit:
\begin{equation}
b_u(S_t)=
\begin{cases}
1, & u\in S_t,\\
\eta_c, &
u\in\bigcup_{i\in S_t}\mathcal M(i)\setminus S_t,\\
0, & \text{otherwise},
\end{cases}
\label{eq:coverage-credit}
\end{equation}
where $u$ indexes target locations and $\eta_c=0.25$.
The coverage score is $C(S_t)=\sum_{u=1}^{N_t}p_u b_u(S_t)$.
Each location contributes once, including when selected
neighborhoods overlap. The local-relevance term
$D(S_t)=\sum_{i\in S_t}d_i$ rewards selected tokens with
high relevance and strong contrast with neighboring
relevance values, where
\begin{equation}
d_i=\alpha_d\left(
p_i+\frac{1}{|\mathcal M(i)|}
\sum_{u\in\mathcal M(i)}|p_u-p_i|
\right).
\label{eq:local-relevance}
\end{equation}
where $d_i$ is the score of target token $i$ and
$\alpha_d=0.25$ weights this contribution.
Jointly, $C$ and $D$ value target evidence independently
of whether a strong reference match is available.

\paragraph{Bilateral support.}
The bilateral term rewards retaining target regions
together with their corresponding reference regions.
Let $\mathcal N(i)$ denote the reference candidate
neighborhood for target token $i$.
Given the soft correspondence weights $a_{ij}$ from the
matching module, the relative match strength is
\begin{equation}
q_{ij}=
\frac{a_{ij}}
{\max_{k\in\mathcal N(i)}a_{ik}},
\qquad j\in\mathcal N(i),
\label{eq:relative-match}
\end{equation}
where $j$ and $k$ index reference tokens.
The strongest candidate match therefore has a score of $1$.
Let $c_i\in[0,1]$ denote matching concentration.
A larger $c_i$ indicates a more concentrated match among
candidate reference tokens. Bilateral support is
\begin{equation}
B(S_t,S_r)=
\lambda_r\sum_{i\in S_t}p_i c_i
\max_{j\in S_r\cap\mathcal N(i)}q_{ij},
\label{eq:bilateral-utility}
\end{equation}
where $\lambda_r=1$ weights the bilateral contribution.
The maximum is zero when no corresponding reference
token is retained.
Each selected target receives credit for its strongest
retained reference, weighted by target relevance and
matching concentration.
An additional reference increases this credit only if
it improves the best retained match.
A reference token can support multiple selected targets.

Thus, target tokens contribute coverage and local relevance,
while reference tokens contribute through their support
for selected targets.
Either type of addition can increase bilateral support
by completing or improving a retained correspondence.

\paragraph{Allocating the budget.}
We initialize selection with a small set of spatially
distributed target anchors.
At each iteration, we consider individual target tokens,
individual reference tokens, and matched target--reference
pairs.
For an addition $A=(A_t,A_r)$, let $A_t$ and $A_r$ contain
the newly added target and reference indices.
Its objective gain is
\begin{equation}
\Delta J(A)=
J(S_t\cup A_t,S_r\cup A_r)-J(S_t,S_r).
\label{eq:selection-gain}
\end{equation}
We greedily choose
\begin{equation}
A^\star=
\arg\max_{A\in\mathcal Q(S_t,S_r)}
\frac{\Delta J(A)}{|A_t|+|A_r|},
\label{eq:greedy-selection}
\end{equation}
where $\mathcal Q(S_t,S_r)$ contains valid additions
that fit the remaining budget.
The denominator counts newly added tokens, giving a cost
of one for a singleton and two for a pair.
Pair candidates allow two-token additions to be evaluated
jointly within the same procedure. After each addition, we update the selected sets and
recompute the gains.
Selection continues until the budget is filled.

\paragraph{Supervised evidence learning.}
The decoder learns to generate lesion type, available visual
attributes, and diagnostic assessment from the selected
target and reference tokens. Lesion annotations supervise
the target relevance map used for selection. The training
objective combines answer and localization losses:
\begin{equation}
\mathcal L_{\mathrm{SFT}}
=\mathcal L_{\mathrm{ans}}
+\lambda_{\mathrm{loc}}\mathcal L_{\mathrm{loc}},
\label{eq:sft}
\end{equation}
where $\lambda_{\mathrm{loc}}$ weights localization
supervision. The answer loss $\mathcal L_{\mathrm{ans}}$
uses field-normalized cross-entropy to supervise the
structured answer, including its format.
The localization loss is
$\mathcal L_{\mathrm{loc}}
=-\sum_{i=1}^{N_t}q_i^{\mathrm{loc}}\log p_i$,
where $p_i$ is the predicted relevance of target token $i$
and $q_i^{\mathrm{loc}}$ is its normalized lesion-overlap
weight derived from the annotations.

During SFT, we jointly optimize the visual encoder and
merger, bilateral enhancement module, relevance head,
and decoder LoRA parameters, while keeping the original
decoder weights frozen. The model predicts lesion attributes
before generating the final diagnostic assessment.
Lesion annotations guide training but are not required
at inference.

\subsection{Bilateral Diagnostic-Support Optimization}
\label{sec:grpo}

Post-training has two goals: improve the correctness of the generated descriptors and assessment, and encourage assessment supported by the retained bilateral context. The visual pathway is frozen while decoder LoRA parameters are updated.

\paragraph{Rewarding lesion characterization and assessment.}
For each input, we sample a group of responses. A base reward scores
diagnosis, lesion type, and available attributes:
\begin{equation}
R_{\mathrm{base}}(y)
=w_d r_{\mathrm{diag}}(y)
+w_t r_{\mathrm{type}}(y)
+w_a r_{\mathrm{attr}}(y).
\label{eq:base-reward}
\end{equation}
Each term checks the corresponding labels against annotations,
giving lesion characterization an explicit training signal alongside
assessment. Free-form explanations do not receive an independent
correctness reward.

\paragraph{Measuring support from the reference.}
The bilateral reward compares diagnostic support with intact and
disrupted reference features. We first select target tokens $S_t$ and reference tokens $S_r$
from the original image pair. We then create $M$ shuffled
versions of the reference features by rearranging spatial
blocks within the same reference image. For each version,
we recompute correspondence and bilateral enhancement
using the unchanged raw target features. We reuse the
originally selected token indices, keeping the token count,
attention masks, and positional indices unchanged.
The resulting representations are $Z_B$ for the intact pair and
$Z_{C_m}$ for control $m$.

For each sampled response $y$, we construct a prefix $r_y$ from its
predicted attribute labels, excluding the diagnosis and free-form
explanation. We use the same prefix and fixed rollout policy for all
intact/control comparisons. Let $s(d\mid Z,r_y)$ be the normalized
log support for diagnosis $d$, computed from the length-averaged
label log-likelihoods of \emph{benign} and \emph{malignant}. For the
ground-truth diagnosis $d^*$, the support difference is
\begin{equation}
\Delta_y=s(d^*\mid Z_B,r_y)
-\frac{1}{M}\sum_{m=1}^{M}s(d^*\mid Z_{C_m},r_y).
\label{eq:bilateral-support-difference}
\end{equation}
A positive $\Delta_y$ means that the intact reference provides
stronger diagnostic support given the same predicted attributes.
We add a bounded positive bonus only when the sampled response is
valid and its diagnosis is correct:
\begin{equation}
\begin{aligned}
R_{\mathrm{bil}}(y)&=v_y\,\operatorname{clip}(\Delta_y,0,\mu),\\
R(y)&=R_{\mathrm{base}}(y)
+\lambda_{\mathrm{bil}}R_{\mathrm{bil}}(y),
\end{aligned}
\label{eq:besr-reward}
\end{equation}
where $v_y$ is 1 for a valid response with a correct diagnosis
and 0 otherwise. The parameters $\mu$ and $\lambda_{\mathrm{bil}}$
control the maximum bonus and its weight, respectively.
We normalize the total rewards within each group of sampled
responses and use them to optimize the policy with
GRPO~\cite{shao2024deepseekmath}.

\paragraph{Inference and efficiency.}
At inference, BESR encodes both mammograms, selects
$K_{\mathrm{eff}}$ visual tokens, and generates one structured
answer. Reference shuffling and diagnostic-support probes
are used only during training. Token selection shortens the
decoder input while retaining the encoding cost of both images.

\section{Experimental Protocol}
\label{sec:experiments}

\begin{table*}[t]
\centering

\caption{\textbf{Token-selection and reasoning-model comparisons.}
(a)~Token selectors at $K{=}256$.
(b)~Medical reasoning models with full input or 256 tokens retained by our selector.
Balanced accuracy (BAcc.), F1, and lesion coverage (Cov.) are in percent,
reported as mean $\pm$ standard deviation over three seeds.
Bold denotes the best result among budget-matched methods.}
\label{tab:comparison}
\small
\setlength{\tabcolsep}{4pt}
\renewcommand{\arraystretch}{1.08}

\begin{tabular*}{\textwidth}{@{\extracolsep{\fill}}lrrrrrr@{}}
\toprule
\multicolumn{7}{l}{\textbf{(a) Token-selection comparison}}\\
\midrule
& \multicolumn{3}{c}{Bilateral CBIS-DDSM}
& \multicolumn{3}{c}{VinDr-Mammo}\\
\cmidrule(lr){2-4}\cmidrule(l){5-7}
Input / selector
& BAcc.$\uparrow$ & F1$\uparrow$ & Cov.$\uparrow$
& BAcc.$\uparrow$ & F1$\uparrow$ & Cov.$\uparrow$\\
\midrule
Full tokens
& $63.60{\scriptstyle\pm 2.04}$ & $59.06{\scriptstyle\pm 2.11}$ & $100{\scriptstyle\pm 0.00}$ & $61.48{\scriptstyle\pm 3.00}$ & $50.00{\scriptstyle\pm 3.17}$ & $100{\scriptstyle\pm 0.00}$\\
GT lesion
& $66.26{\scriptstyle\pm 0.88}$ & $61.71{\scriptstyle\pm 1.55}$ & $100{\scriptstyle\pm 0.00}$ & $64.78{\scriptstyle\pm 1.67}$ & $60.87{\scriptstyle\pm 1.42}$ & $100{\scriptstyle\pm 0.00}$\\
\midrule
\multicolumn{7}{l}{\textit{Budget-matched selection ($K=256$)}}\\
\midrule
Spatial Sampling
& $61.21{\scriptstyle\pm 2.53}$ & $61.25{\scriptstyle\pm 2.62}$ & $11.37{\scriptstyle\pm 0.00}$ & $64.00{\scriptstyle\pm 1.40}$ & $57.27{\scriptstyle\pm 3.58}$ & $12.49{\scriptstyle\pm 0.00}$\\
MMTok~\cite{dong2026mmtok}
& $65.15{\scriptstyle\pm 1.83}$ & $61.88{\scriptstyle\pm 3.22}$ & $67.84{\scriptstyle\pm 5.99}$ & $64.05{\scriptstyle\pm 1.16}$ & $53.33{\scriptstyle\pm 4.18}$ & $41.97{\scriptstyle\pm 5.92}$\\
DivPrune~\cite{alvar2025divprune}
& $62.07{\scriptstyle\pm 1.15}$ & $63.25{\scriptstyle\pm 3.13}$ & $22.18{\scriptstyle\pm 4.37}$ & $63.67{\scriptstyle\pm 2.65}$ & $57.01{\scriptstyle\pm 5.74}$ & $9.78{\scriptstyle\pm 2.08}$\\
VisionZip~\cite{yang2025visionzip}
& $65.61{\scriptstyle\pm 2.74}$ & $58.46{\scriptstyle\pm 4.37}$ & $60.62{\scriptstyle\pm 2.82}$ & $60.89{\scriptstyle\pm 1.36}$ & $59.00{\scriptstyle\pm 3.72}$ & $53.47{\scriptstyle\pm 2.88}$\\
CROP-ILP~\cite{guo2025crop}
& $65.52{\scriptstyle\pm 1.29}$ & $58.79{\scriptstyle\pm 3.05}$ & $55.33{\scriptstyle\pm 0.00}$ & $58.18{\scriptstyle\pm 3.39}$ & $58.95{\scriptstyle\pm 6.75}$ & $37.27{\scriptstyle\pm 0.00}$\\
\textbf{Bilateral selector (ours)}
& $\mathbf{71.30}{\scriptstyle\pm\mathbf{2.11}}$ & $\mathbf{68.77}{\scriptstyle\pm\mathbf{2.42}}$ & $\mathbf{96.76}{\scriptstyle\pm\mathbf{0.47}}$ & $\mathbf{66.84}{\scriptstyle\pm\mathbf{1.43}}$ & $\mathbf{62.39}{\scriptstyle\pm\mathbf{2.35}}$ & $\mathbf{93.91}{\scriptstyle\pm\mathbf{0.67}}$\\
\bottomrule
\end{tabular*}

\vspace{0.7em}

\begin{tabular*}{\textwidth}{@{\extracolsep{\fill}}llrrrr@{}}
\toprule
\multicolumn{6}{l}{\textbf{(b) Comparison with reasoning models}}\\
\midrule
& & \multicolumn{2}{c}{Bilateral CBIS-DDSM}
& \multicolumn{2}{c}{VinDr-Mammo}\\
\cmidrule(lr){3-4}\cmidrule(l){5-6}
Model & Input
& BAcc.$\uparrow$ & F1$\uparrow$
& BAcc.$\uparrow$ & F1$\uparrow$\\
\midrule
VisCoT~\cite{shao2024visualcot}
& Full
& $57.99{\scriptstyle\pm 0.62}$ & $62.40{\scriptstyle\pm 0.61}$ & $50.99{\scriptstyle\pm 0.80}$ & $61.86{\scriptstyle\pm 0.48}$\\
Med-R1~\cite{lai2026medr1}
& Full
& $54.79{\scriptstyle\pm 1.72}$ & $37.84{\scriptstyle\pm 3.49}$ & $65.87{\scriptstyle\pm 2.75}$ & $56.00{\scriptstyle\pm 3.45}$\\
MedReason-R1~\cite{li2026medreasonr1}
& Full
& $60.41{\scriptstyle\pm 0.56}$ & $55.06{\scriptstyle\pm 0.44}$ & $60.54{\scriptstyle\pm 4.17}$ & $40.32{\scriptstyle\pm 5.19}$\\
MedVLThinker~\cite{huang2026medvlthinker}
& Full
& $57.86{\scriptstyle\pm 1.17}$ & $51.11{\scriptstyle\pm 1.33}$ & $63.48{\scriptstyle\pm 1.44}$ & $48.12{\scriptstyle\pm 6.75}$\\
ViTAR~\cite{chen2025think}
& Full
& $49.90{\scriptstyle\pm 0.90}$ & $54.51{\scriptstyle\pm 0.72}$ & $61.85{\scriptstyle\pm 2.88}$ & $52.23{\scriptstyle\pm 3.78}$\\
\midrule
Med-R1
& 256 tokens
& $65.00{\scriptstyle\pm 0.49}$ & $57.32{\scriptstyle\pm 0.64}$ & $66.09{\scriptstyle\pm 2.57}$ & $58.00{\scriptstyle\pm 4.15}$\\
MedReason-R1
& 256 tokens
& $64.50{\scriptstyle\pm 1.01}$ & $55.08{\scriptstyle\pm 1.29}$ & $65.00{\scriptstyle\pm 2.56}$ & $57.14{\scriptstyle\pm 2.01}$\\
MedVLThinker
& 256 tokens
& $65.64{\scriptstyle\pm 0.45}$ & $57.86{\scriptstyle\pm 0.62}$ & $62.72{\scriptstyle\pm 2.40}$ & $56.65{\scriptstyle\pm 2.24}$\\
ViTAR
& 256 tokens
& $65.41{\scriptstyle\pm 0.62}$ & $62.03{\scriptstyle\pm 0.61}$ & $61.30{\scriptstyle\pm 2.88}$ & $51.28{\scriptstyle\pm 3.78}$\\
\midrule
\textbf{BESR (ours)}
& 256 tokens
& $\mathbf{73.73}{\scriptstyle\pm\mathbf{1.01}}$ & $\mathbf{73.00}{\scriptstyle\pm\mathbf{1.84}}$ & $\mathbf{70.85}{\scriptstyle\pm\mathbf{2.32}}$ & $\mathbf{68.27}{\scriptstyle\pm\mathbf{1.35}}$\\
\bottomrule
\end{tabular*}
\end{table*}

\paragraph{Datasets.}
We evaluate on bilateral CBIS-DDSM for pathology-confirmed
benign--malignant classification and VinDr-Mammo for binary
BI-RADS assessment prediction. Starting from each dataset's
training and test splits, we pair opposite breasts from the
same examination and projection. The contralateral breast
provides a patient-specific reference and is not assumed
normal. For CBIS-DDSM~\cite{lee2017cbis}, we link lesion records
to original DDSM examinations~\cite{heath1998current}; lesion
annotations and mass and calcification descriptors supervise
evidence learning. For VinDr-Mammo~\cite{nguyen2023vindr},
the target breast's BI-RADS defines low suspicion (1--3)
versus high suspicion (4--5), rather than pathology-confirmed
malignancy. The final single-target cohorts contain 1,983
and 1,299 examples, respectively. We train first on bilateral
CBIS-DDSM and then fine-tune on VinDr-Mammo for its
BI-RADS assessment task. Split corrections, validation
construction, eligibility criteria, and exact cohort statistics
are provided in the supplement.

\paragraph{Baselines and ablations.}
We use Qwen2.5-VL-7B and evaluate visual-token budgets
$K\in\{64,128,256\}$. For evidence selection, we compare
our bilateral selector with spatial sampling and existing
token-selection methods under matched supervised training
and evaluation settings. Full-token and ground-truth lesion
inputs serve as additional references. We also ablate
bilateral enhancement and joint target--reference selection.

To evaluate evidence utilization, we compare SFT and
GRPO with and without the bilateral reward, keeping
the visual pathway fixed during post-training.
We also test intact, shuffled, and disabled contralateral
references and compare BESR with medical reasoning
models using full inputs or 256 selected tokens.

\begin{figure}[t]
    \centering
    \includegraphics[width=\columnwidth]{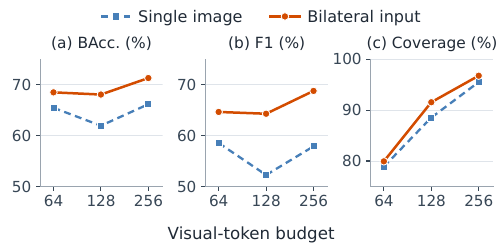}
    \caption{\textbf{Single-image versus bilateral selection on
    CBIS-DDSM.} Balanced accuracy, F1, and lesion coverage are
    compared across visual-token budgets $K\in\{64,128,256\}$.}
    \label{fig:single_bilateral}
\end{figure}

\paragraph{Evaluation.}
We report balanced accuracy and positive-class F1, treating
malignant as positive for CBIS-DDSM and high suspicion
(BI-RADS 4--5) as positive for VinDr-Mammo. Lesion coverage
measures the fraction of annotated lesion pixels covered by
retained token footprints, while descriptor accuracy evaluates
the correctness of predicted lesion attributes. Together,
these metrics assess diagnostic performance, spatial evidence
retention, and lesion characterization. We repeat experiments
with three random training seeds and report the mean and sample
standard deviation across runs, keeping the training,
validation, and test splits fixed. We evaluate inference
efficiency using prefill and end-to-end latency.

\begin{figure}[t]
\centering
\includegraphics[width=0.5\textwidth]{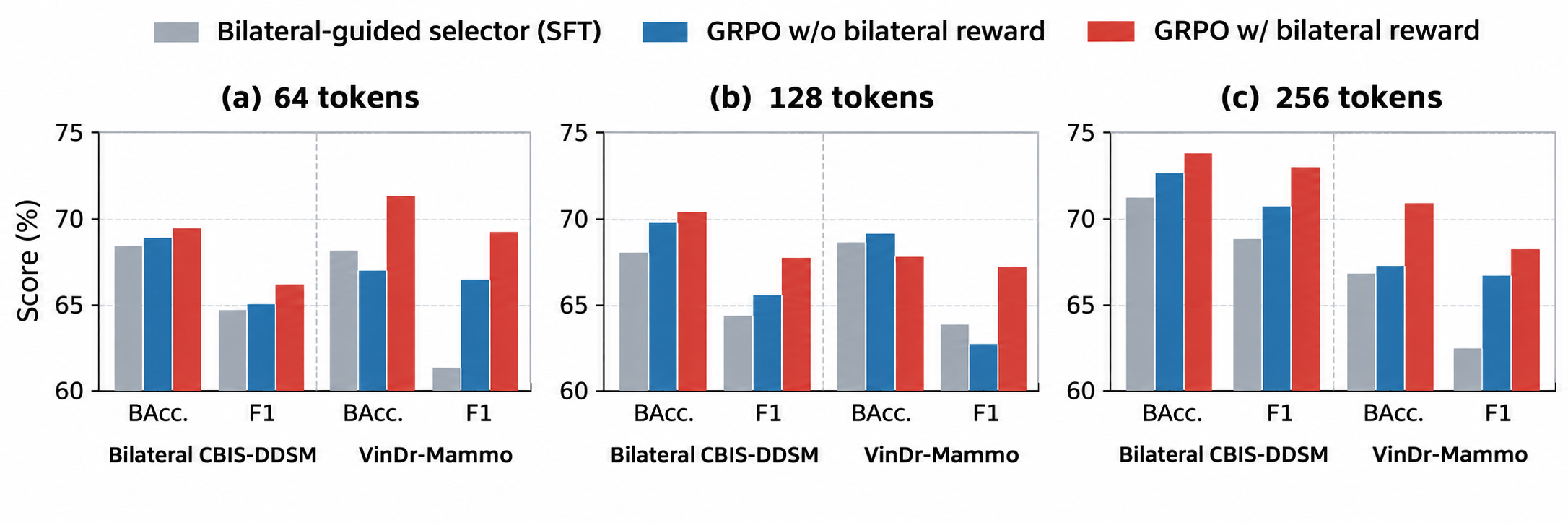}
\caption{\textbf{Effect of GRPO post-training.}
Balanced accuracy and F1 (\%) on bilateral CBIS-DDSM and
VinDr-Mammo at (a) $K=64$, (b) $K=128$, and (c) $K=256$.
Gray, blue, and red represent supervised fine-tuning (SFT),
GRPO without the bilateral reward, and GRPO with the
bilateral reward, respectively.}
\label{fig:grpo}
\end{figure}

\begin{figure}[t]
\centering
\includegraphics[width=\columnwidth]{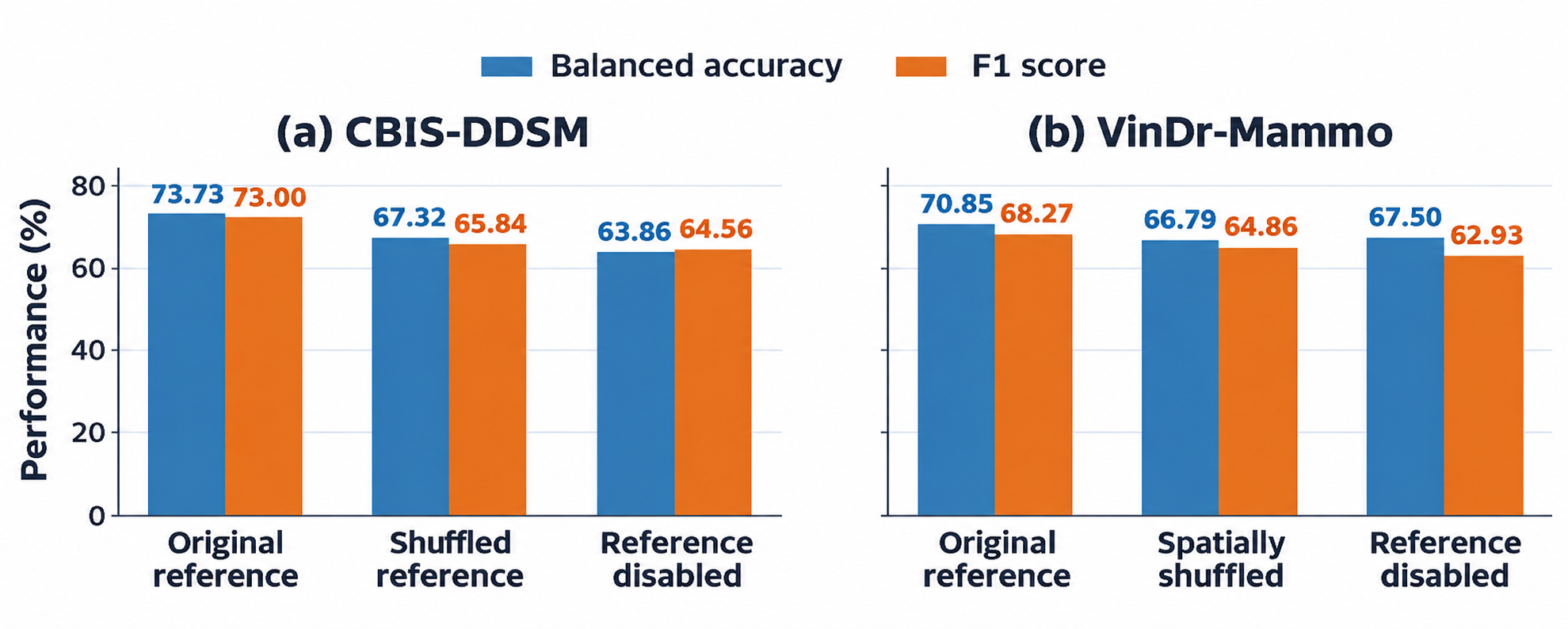}
\caption{\textbf{Effect of the contralateral reference.} Balanced accuracy and F1 on CBIS-DDSM and VinDr-Mammo under intact, shuffled, and disabled reference conditions
at $K=256$}
\label{fig:single_bilateral}
\end{figure}

\begin{figure}[t]
\centering
\includegraphics[width=0.45\textwidth]{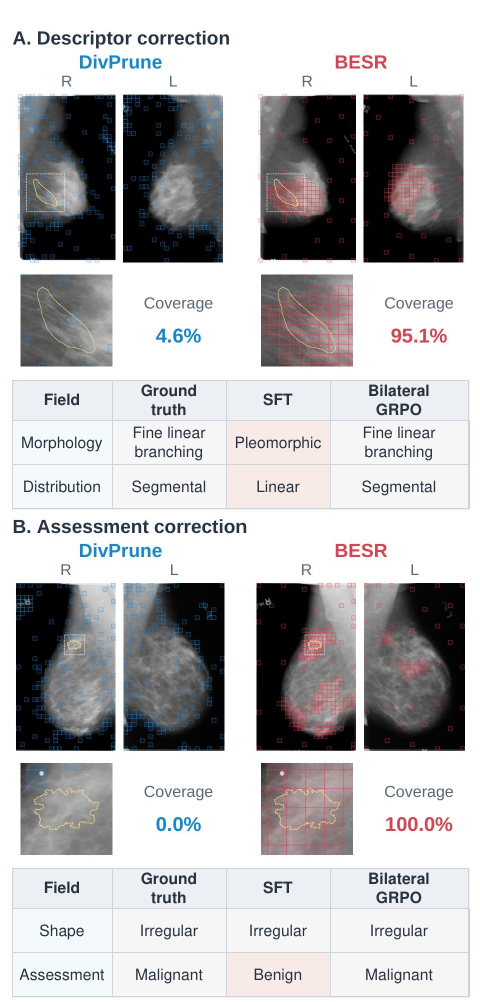}
\caption{\textbf{Evidence retention and interpretation.}
At $K=256$ tokens per bilateral pair, Case A illustrates
descriptor correction and Case B illustrates assessment
correction. Each case compares token selections from
separately fine-tuned DivPrune and BESR models, followed
by SFT and bilateral GRPO predictions using identical
BESR-selected tokens. Yellow contours indicate ground-truth
lesion annotations for visualization only. Pink cells
highlight predictions that differ from ground truth.}
\label{fig:grpo}
\end{figure}

\section{Results and Analysis}
\label{sec:results}

\subsection{Evidence Selection and Retention}
Table~1(a) shows that our bilateral selector achieves the best
balanced accuracy, F1, and annotated-lesion coverage among the
evaluated budget-matched selectors on both datasets. On CBIS-DDSM,
it improves balanced accuracy by 5.69 percentage points over the
strongest competing selector at $K=256$. Full-token and ground-truth
lesion inputs provide complete lesion coverage but lower predictive
performance, indicating that covering a lesion is not sufficient
for accurate assessment. Figure~3 further shows that bilateral selection consistently improves
over single-image selection across token budgets.
Multi-method comparisons and selection visualizations are
provided in Figs.~S1 and S3.

\paragraph{Complementary benefits of bilateral enhancement and joint selection.}
Table~\ref{tab:selector_ablation} separates two ways of using the
contralateral breast. First, bilateral enhancement improves balanced
accuracy from 66.16\% to 68.72\% while all 256 retained tokens still
come from the target breast. Thus, reference information can improve
target features without explicitly retaining reference tokens.
Second, allocating the same budget jointly across both breasts
further increases balanced accuracy to 71.30\%, although lesion
coverage decreases slightly relative to enhancement alone.
Together, these findings suggest that retaining complementary
reference evidence provides information beyond target lesion.

\subsection{Effect of Bilateral GRPO}

Preserving visual evidence does not guarantee accurate
diagnostic interpretation. Figure~4 compares SFT and GRPO
with and without the bilateral reward, keeping the visual
pathway fixed during post-training. At $K=256$, GRPO with
the bilateral reward improves CBIS-DDSM balanced accuracy
and F1 over SFT by 2.43 and 4.23 percentage points,
respectively, and outperforms GRPO without the reward on
both datasets. These results suggest an additional benefit
from bilateral diagnostic support beyond standard GRPO.
However, the gains are not uniform: on VinDr-Mammo at
$K=128$, F1 improves while balanced accuracy decreases.
Supplementary Fig.~S2 further shows improvements in
several lesion attributes after Bilateral GRPO.

Figure~5 provides complementary evidence about reference use.
Shuffling or disabling the contralateral reference reduces both
balanced accuracy and F1 on the two datasets. The results show that
predictions benefit from intact bilateral information, although
sensitivity to reference disruption alone does not establish
anatomically faithful reasoning.

\begin{table}[t]
\centering
\small
\caption{\textbf{Selector ablation at $K=256$ on CBIS.}
Components are added cumulatively. Target only and bilateral
enhancement retain all 256 tokens from the target image.
Joint selection allocates the same total budget across the
target and reference images.}
\label{tab:selector_ablation}
\setlength{\tabcolsep}{2pt}
\renewcommand{\arraystretch}{1.15}
\begin{tabular*}{\columnwidth}{@{\extracolsep{\fill}}lrrr@{}}
\hline
\noalign{\vskip 2pt}
Configuration & BAcc. (\%) & F1 (\%) & Coverage (\%) \\[2pt]
\hline
\noalign{\vskip 2pt}
Target only
& $66.16\,{\scriptstyle\pm 0.88}$
& $57.88\,{\scriptstyle\pm 1.55}$
& $95.52\,{\scriptstyle\pm 0.60}$  \\
+ Bilateral enhancement
& $68.72\,{\scriptstyle\pm 1.02}$
& $67.48\,{\scriptstyle\pm 1.35}$
& $97.23\,{\scriptstyle\pm 0.59}$ \\
+ Joint selection
& $\mathbf{71.30}\,{\scriptstyle\pm 2.11}$
& $\mathbf{68.77}\,{\scriptstyle\pm 2.42}$
& $\mathbf{96.76}\,{\scriptstyle\pm 0.47}$ \\[2pt]
\hline
\end{tabular*}
\end{table}

\begin{table}[t]
\centering
\caption{\textbf{Inference efficiency across token budgets.}
Token selection reduces prefill latency, with smaller end-to-end
gains and nearly unchanged peak GPU memory.
Full-token inference uses 1,979 visual tokens on average.}
\label{tab:inference_efficiency}
\small
\setlength{\tabcolsep}{3pt}
\begin{tabular*}{\columnwidth}{@{\extracolsep{\fill}}lrrrr@{}}
\toprule
Method & Tokens & Prefill & End-to-end & Memory \\
& & (ms) & (s) & (GiB) \\
\midrule
Full tokens   & 1,979 & 183.16 & 1.780 & 18.198 \\
Ours, $K=256$ &   256 &  40.50 & 1.595 & 18.095 \\
Ours, $K=128$ &   128 &  35.86 & 1.576 & 18.087 \\
Ours, $K=64$  &    64 & \textbf{27.70} & \textbf{1.563} & 18.083 \\
\bottomrule
\end{tabular*}
\end{table}
\subsection{Comparison with Medical Reasoning Models}
Table~1(b) compares BESR with medical reasoning models using either
full visual inputs or 256 tokens retained by our supervised bilateral
selector. BESR achieves the highest balanced accuracy and F1 on both
datasets. On CBIS-DDSM, it exceeds the strongest 256-token baselines
by 8.09 and 10.97 percentage points, respectively; the corresponding
gains on VinDr-Mammo are 4.76 and 10.27 points. 

Two findings emerge from these comparisons. First, using
selected tokens generally improves the reasoning baselines
on CBIS-DDSM, although the results on VinDr-Mammo are mixed.
Second, none of the 256-token reasoning models outperforms
the supervised bilateral selector, despite using the same
selected visual evidence. This suggests that retaining
relevant evidence alone is insufficient and learning to
interpret it is equally important. Overall, BESR demonstrates
stronger diagnostic performance, although the compared
approaches use different training methods.

\subsection{Inference Efficiency}
Table~\ref{tab:inference_efficiency} shows that BESR reduces
the decoder's visual input by 87.1\% at $K=256$, achieving
a $4.52\times$ prefill speedup. This demonstrates that
selecting compact bilateral evidence can substantially
reduce decoder computation. End-to-end inference also
improves by $1.12\times$, with smaller gains because
both images still require visual encoding.

\subsection{Qualitative Evidence and Interpretation}

Figure~6 illustrates two complementary benefits of BESR.
The bilateral selector improves lesion evidence retention
under a limited token budget, while bilateral GRPO helps
interpret the retained evidence for lesion characterization
and diagnostic assessment. In Case~A, bilateral GRPO
corrects the lesion descriptors using the same selected
visual evidence as SFT. 
In Case~B, SFT correctly identifies the irregular lesion
shape but predicts a benign assessment, whereas bilateral
GRPO produces the correct malignant assessment using
the same retained visual evidence.
 Additional qualitative examples are
provided in supplementary Fig.~S4.

\section{Conclusion}
\label{sec:conclusion}

We presented Bilateral Evidence Selection and Reasoning (BESR)
to address two challenges in efficient medical vision--language
learning: preserving diagnostic evidence during token compression
and learning to interpret it for assessment. Experiments on
bilateral CBIS-DDSM and VinDr-Mammo demonstrate improved evidence
retention and diagnostic performance under limited token budgets,
highlighting the importance of preserving both lesion evidence
and its bilateral context. Ablation studies show the complementary
benefits of bilateral enhancement and joint selection, while
the bilateral reward further improves diagnostic assessment.
BESR also substantially reduces decoder prefill latency,
demonstrating the efficiency of selective evidence retention.
Future work will extend BESR to examinations with multiple
findings and evaluate its generalization across clinical settings.

\section{Acknowledgments}
This work was supported in part by a NIH Other Transaction research contract \#1OT2OD037972-01 and \#3OT2OD037972-01S1, the grant 1R01EB032896 as part of the NSF/NIH Smart Health and Biomedical Research in the Era of Artificial Intelligence and Advanced Data Science Program, an Amazon Machine Learning Research Award, a PA Breast Cancer Coalition grant, a Jewish Healthcare Foundation grant, and the University of Pittsburgh Momentum Funds (Scaling Grant) for the Pittsburgh Center for AI Innovation in Medical Imaging. This work used Bridges-2 at the Pittsburgh Supercomputing Center through allocation [MED200006] from the Advanced Cyberinfrastructure Coordination Ecosystem: Services \& Support (ACCESS) program, which is supported by NSF grants \#2138259, \#2138286, \#2138307, \#2137603, and \#2138296. This research was also supported in computing resources by the University of Pittsburgh Center for Research Computing and Data (RRID: SCR\_022735) through the resources provided by the H2P cluster, which is supported by NSF award OAC-2117681. The views and conclusions contained in this document are those of the authors and should not be interpreted as representing official policies, either expressed or implied, of the NIH or NSF.

\clearpage
\appendix
\setcounter{figure}{0}
\setcounter{table}{0}
\setcounter{equation}{0}

\renewcommand{\thefigure}{S\arabic{figure}}
\renewcommand{\thetable}{S\arabic{table}}
\renewcommand{\theequation}{S\arabic{equation}}

\section*{Supplementary Material}

This supplement expands the joint token selector, supervised training,
bilateral diagnostic-support reward, and evaluation protocol described in
the main paper. It also specifies the structured response format and
provides additional analyses of token budgets, retained evidence, lesion
attributes, and predictions before and after post-training.

\section{Cohort Composition and Prediction Targets}
\label{sec:supp_cohorts}

\paragraph{Ethics and Data Use.}
We use two publicly available, de-identified datasets.
For CBIS-DDSM, patient consent was obtained during
the original DDSM collection~\cite{lee2017cbis}.
VinDr-Mammo received institutional review board approval,
with informed consent waived~\cite{nguyen2023vindr}.

Table S1 summarizes the single-target cohorts, with each example pairing a target mammogram and a contralateral image from the same examination and projection. CBIS-DDSM uses pathology-based benign–malignant labels. VinDr-Mammo uses the target breast’s BI-RADS assessment, grouped into low suspicion (1–3) and high suspicion (4–5); these labels do not establish confirmed malignancy.

\begin{table}[t]
\centering
\small
\caption{\textbf{Dataset composition by split.} Counts denote input examples.}
\label{tab:supp_cohort_counts}
\setlength{\tabcolsep}{3pt}
\renewcommand{\arraystretch}{1.08}
\begin{tabular*}{\columnwidth}{@{\extracolsep{\fill}}llrrr@{}}
\hline
\noalign{\vskip 2pt}
\textbf{Dataset} & \textbf{Class} & \textbf{Train} & \textbf{Val.} & \textbf{Test} \\[2pt]
\hline
\noalign{\vskip 2pt}
CBIS-DDSM & Benign    & 759 & 60 & 232 \\
          & Malignant & 672 & 80 & 180 \\
          & \textbf{Total} & \textbf{1,431} & \textbf{140} & \textbf{412} \\[2pt]
\hline
\noalign{\vskip 2pt}
VinDr-Mammo & BI-RADS 1--3 & 507 & 58 & 151 \\
            & BI-RADS 4--5 & 424 & 39 & 120 \\
            & \textbf{Total} & \textbf{931} & \textbf{97} & \textbf{271} \\[2pt]
\hline
\end{tabular*}
\end{table}

\section{Inputs and Structured Predictions}
\label{sec:supp-inputs}

Each input pairs opposite breasts from the same examination and
projection (CC/CC or MLO/MLO). Image orientation, lesion coordinates,
and annotations undergo consistent preprocessing. The opposite breast
is not assumed normal. The shared budget $K_{\mathrm{eff}}$ counts only
tokens passed to the decoder, not those processed by the vision encoder.

The response describes one target finding. Masses use shape and margin,
while calcifications use morphology and distribution. Applicable
attributes precede the final assessment so that diagnostic-support
probes can condition on the model's own predicted attributes.
Ground-truth labels and lesion annotations are not given at inference.
Unannotated attributes are omitted from the corresponding losses and
rewards, and explanations receive no independent correctness reward.
The expected answer formats are:

\noindent\textbf{Mass response.}
\begin{lstlisting}[style=vlmprompt]
{
  "lesion_type": "mass",
  "shape": "<predicted shape>",
  "margin": "<predicted margin>",
  "explanation": "<concise explanation>",
  "diagnosis": "<benign or malignant>"
}
\end{lstlisting}

\noindent\textbf{Calcification response.}
\begin{lstlisting}[style=vlmprompt]
{
  "lesion_type": "calcification",
  "morphology": "<predicted morphology>",
  "distribution": "<predicted distribution>",
  "explanation": "<concise explanation>",
  "diagnosis": "<benign or malignant>"
}
\end{lstlisting}

\section{Bilateral Selection: Implementation Details}
\label{sec:supp_selection}
The bilateral enhancement, joint selection objective, and greedy budget
allocation are defined in main-paper Eqs.~(1)--(9). Here we specify
correspondence parameters and implementation choices omitted there.

\subsection{Correspondence and Enhancement}
For normalized target/reference coordinates $x_i^t,x_j^r$, candidate
reference tokens are drawn from a spatial neighborhood
\begin{equation}
\mathcal N(i)=\{j:\lVert x_i^t-x_j^r\rVert_\infty\leq r\}.
\end{equation}
where $r=0.25$ is the spatial matching radius. Combining original and projected feature similarities with a spatial
penalty gives
\begin{align}
 e_{ij}&=\frac{\cos(h_i^t,h_j^r)+\tanh(\gamma)\cos(W_qh_i^t,W_kh_j^r)}{\tau}
   \notag\\[-2pt]
 &\qquad-\frac{\lVert x_i^t-x_j^r\rVert_2^2}{2r^2},\\
 a_{ij}&=\frac{\exp(e_{ij})}{\sum_{k\in\mathcal N(i)}\exp(e_{ik})},\quad
 j\in\mathcal N(i).
\end{align}
Here $W_q,W_k$ are learned projections, $\gamma$ controls the projected
similarity contribution, and $\tau$ is the matching temperature.
Matching concentration is obtained from normalized attention entropy:
\begin{equation}
c_i=1+\frac{\sum_{j\in\mathcal N(i)}a_{ij}\log a_{ij}}
{\log|\mathcal N(i)|}.
\end{equation}
We compute the match-concentration score for neighborhoods
containing at least two reference candidates. A higher
score indicates that the correspondence weights are
concentrated on fewer candidates. Using the pooled reference feature
$\bar h_i^r$ defined in main-paper Eq.~(1), enhancement uses
\begin{align}
u_i&=\operatorname{GELU}\!\left(W_h\operatorname{LN}
[h_i^t;\bar h_i^r;h_i^t-\bar h_i^r]\right),\\
g_i&=\sigma\!\left(w_g^\top[u_i;c_i]+b_g\right).
\end{align}
The residual update in main-paper Eq.~(1) uses $\psi=W_ou_i$, where $W_o$ projects
the fused features back to the original feature dimension.
We zero-initialize $W_o$ and set $g_i=0$ for invalid matches.

\section{Supervised Training: Additional Details}
\label{sec:supp_sft}
The supervised objective is defined in main-paper Eq.~(10). The
field-normalized answer loss uses
\begin{equation}
\mathcal L_{\mathrm{ans}}=\mathcal L_{\mathrm{diag}}+
0.1\mathcal L_{\mathrm{format}}+
0.1\mathcal L_{\mathrm{type+attr}}.
\end{equation}
The last term averages available lesion-type and attribute fields.
For annotated regions $\mathcal A$, let $n_{\ell i}$ denote the
annotation pixels from region $\ell$ in target token $i$.
The localization target is
\begin{equation}
q_i^{\mathrm{loc}}=\frac1{|\mathcal A|}
\sum_{\ell\in\mathcal A}\frac{n_{\ell i}}{\sum_j n_{\ell j}}.
\end{equation}
Regions with nonzero overlap contribute equally to this target.
Bounding boxes in VinDr-Mammo are treated as annotated regions, not
precise lesion contours. The localization weight is 0.1. Table~\ref{tab:supp_sft_config} provides the training settings.

\begin{table*}[t]
\centering
\caption{\textbf{Supervised-stage configuration.} Answer-loss weights are ordered as diagnosis, format, and type/attributes.}
\label{tab:supp_sft_config}
\footnotesize
\setlength{\tabcolsep}{2.5pt}
\renewcommand{\arraystretch}{1.14}
\begin{tabular}{@{}p{0.25\textwidth}p{0.205\textwidth}p{0.265\textwidth}p{0.19\textwidth}@{}}
\toprule
Setting & Value & Setting & Value\\
\midrule
Backbone & Qwen2.5-VL-7B-Instruct & Paired projection & CC/CC or MLO/MLO\\
Pixel cap per image & 802,816 & Total visual budget $K$ & 64 / 128 / 256\\
Matching projection dimension & 64 & Enhancement hidden dimension & 128\\
Matching temperature $\tau$ & 0.1 & Normalized matching radius $r$ & 0.25\\
Target-anchor allocation & Up to $K_{\mathrm{eff}}/16$ & Target/reference allocation & Adaptive\\
Coverage-neighborhood radius & 1 grid cell & Neighbor credit $\eta_{\mathrm{cov}}$ & 0.25\\
Local-relevance weight $\beta_{\mathrm{det}}$ & 0.25 & Pair-utility weight $\lambda_r$ & 1.0\\
Optimizer / learning rate & AdamW / $2\times10^{-5}$ & Weight decay / clipping & 0.01 / 1.0\\
Epochs / effective batch size & 3 / 8  & LoRA rank / scale / dropout & 16 / 16 / 0\\
Answer-loss weights & 1 / 0.1 / 0.1 & Localization weight $\lambda_{\mathrm{loc}}$ & 0.1\\
Checkpoint selection & Validation BAcc. & Development seed & 42\\
\bottomrule
\end{tabular}
\end{table*}

\section{Bilateral Reward: Additional Details}
\label{sec:supp_grpo}
This section provides implementation details for the
bilateral reward defined in main-paper Eqs.~(11)--(13).

\subsection{Reward Handling}
Attribute correctness is averaged over annotated fields
and requires the correct lesion type. Missing attribute
labels are excluded, and the remaining reward weights
are renormalized. Invalid or truncated responses receive
zero base reward.

\subsection{Diagnostic-Support Computation}
For each sampled response $y$, we construct a prefix $r_y$
from its predicted lesion attributes, excluding the diagnosis
and explanation. Let $Z$ denote the visual representation
from either the intact or shuffled reference, and let
$\mathcal D$ be the set of candidate diagnostic labels.

For each label $d\in\mathcal D$, we compute its
length-averaged log likelihood under the fixed rollout
policy $\pi_{\mathrm{old}}$:
\begin{equation}
\ell(d\mid Z,r_y)
=\frac{1}{|d|}\sum_{u=1}^{|d|}
\log\pi_{\mathrm{old}}(d_u\mid Z,r_y,d_{<u}),
\end{equation}
where $|d|$ is the number of tokens in the label,
$d_u$ is its $u$-th token, and $d_{<u}$ denotes
the preceding label tokens.

We normalize the scores across candidate labels:
\begin{equation}
s(d\mid Z,r_y)
=\ell(d\mid Z,r_y)
-\log\sum_{d'\in\mathcal D}
\exp\ell(d'\mid Z,r_y).
\end{equation}
The same predicted-attribute prefix and rollout policy
are used for intact and shuffled references. Examples without valid
reference controls receive only the base reward.

\subsection{GRPO Configuration}
Starting from the SFT checkpoint, GRPO updates only
the decoder attention LoRA parameters. The optimization
settings are summarized in Table~\ref{tab:supp_grpo_config}.
For the comparison without the bilateral reward, we set
$\lambda_{\mathrm{bil}}=0$.

\begin{table}[t]
\centering
\caption{\textbf{Bilateral-GRPO settings} for CBIS-DDSM at $K=256$.
The comparison without the bilateral reward sets $\lambda_{\mathrm{bil}}=0$.}
\label{tab:supp_grpo_config}
\footnotesize
\setlength{\tabcolsep}{3pt}
\begin{tabular}{@{}p{0.53\columnwidth}p{0.39\columnwidth}@{}}
\toprule
Setting & Value\\
\midrule
Initial / KL-reference policy & SFT checkpoint\\
Training examples / epochs & 1,431 / 3\\
Optimizer updates & 1,074\\
Prompts per update / group size $G$ & 4 / 4\\
Sampling temperature & 1.0\\
Maximum response length & 192 tokens\\
Optimizer / learning rate & AdamW / $10^{-6}$\\
Clip $\epsilon_{\mathrm{clip}}$ / KL $\beta_{\mathrm{KL}}$ & 0.2 / 0.02\\
LoRA rank / scale / dropout & 16 / 16 / 0\\
Base weights $w_d/w_t/w_a$ & 0.5 / 0.2 / 0.3\\
Bilateral weight $\lambda_{\mathrm{bil}}$ & 0.1\\
Bonus cap $\mu$ & 1.0\\
Reference controls $M$ & 2\\
Control construction & Block permutation\\
\bottomrule
\end{tabular}
\end{table}

\section{Evaluation: Additional Details}
\label{sec:supp_eval}
\subsection{Metric and Adaptation Details}
The compared models are scored using their final dataset-specific
assessment, regardless of output format. Lesion-attribute accuracy is
calculated only for the available annotations and independently of the
final assessment. For annotated region $A$ and selected target-token
footprint union $U(S_t)$, lesion coverage is
\begin{equation}\label{eq:supp_coverage}
\operatorname{Cov}(A,S_t)=\frac{|A\cap U(S_t)|}{|A|}.
\end{equation}

\section{Additional Results}
\label{sec:supp-qualitative}

\subsection{Token-Selection Performance across Budgets}

\begin{figure}[t]
\centering
\includegraphics[width=0.5\textwidth,height=0.72\textheight,keepaspectratio]{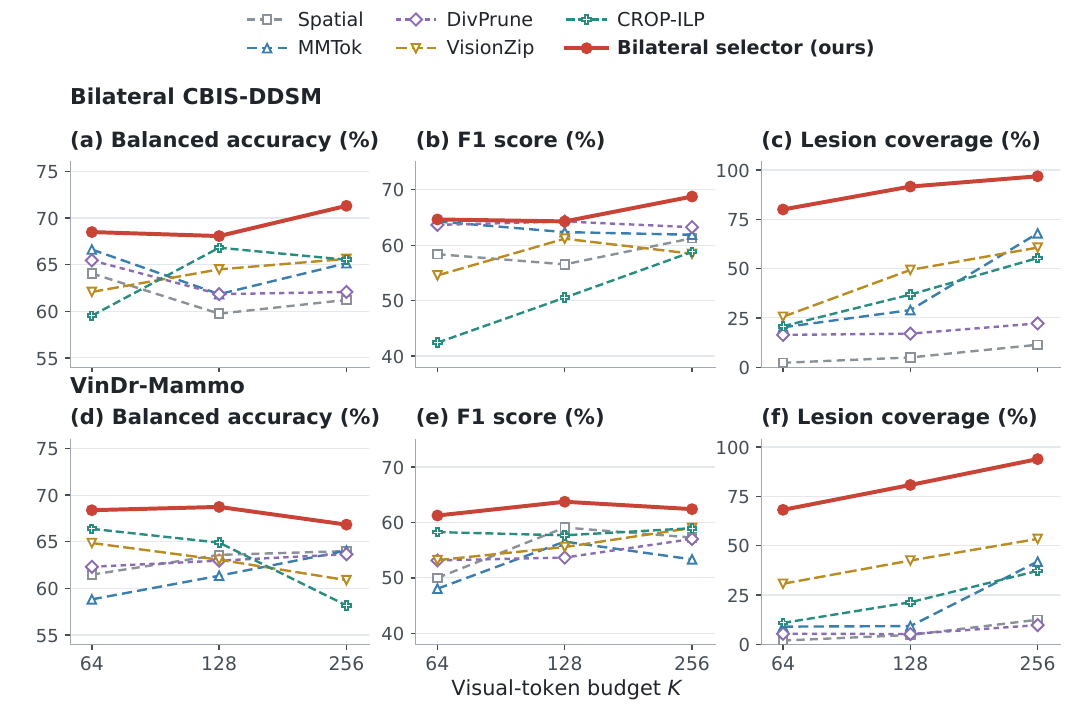}
\caption{\textbf{Token-selection performance across visual-token budgets.}
Balanced accuracy, F1, and lesion coverage on CBIS-DDSM (top) and
VinDr-Mammo (bottom) at $K\in\{64,128,256\}$. VinDr-Mammo coverage is
computed relative to annotated lesion boxes. }
\label{fig:supp-selector-budget}
\end{figure}

Figure~\ref{fig:supp-selector-budget} compares prediction performance
and spatial evidence retention at $K\in\{64,128,256\}$ on both datasets.
On VinDr-Mammo, the bilateral selector's coverage increases with
budget, while balanced accuracy and F1 peak at $K=128$. Increased
annotation coverage therefore does not necessarily yield higher
classification accuracy.

\begin{figure}[t]
\centering
\includegraphics[width=0.5\textwidth,height=0.72\textheight,keepaspectratio]{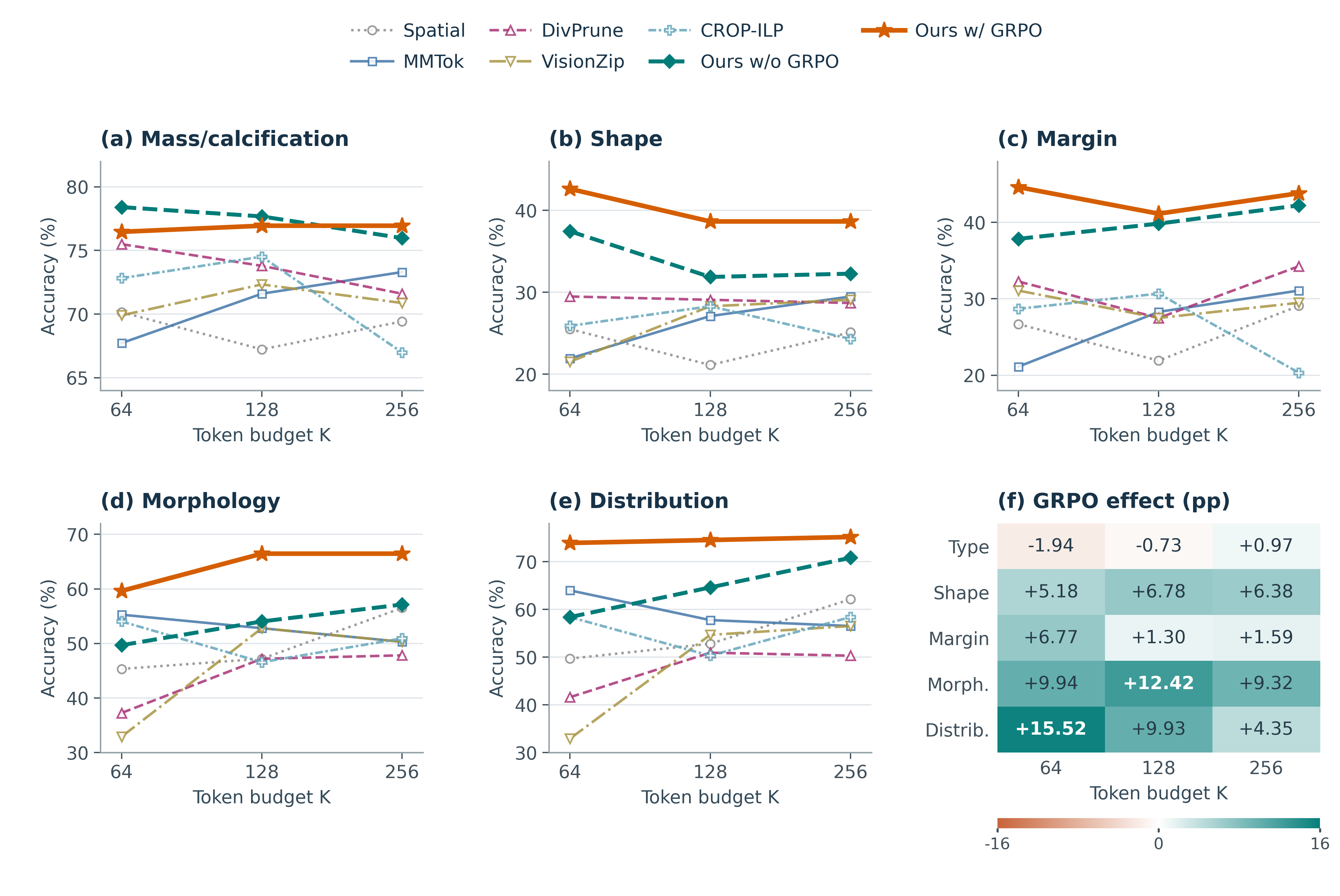}
\caption{\textbf{Lesion-type and attribute prediction across token budgets.}
Panels (a)--(e) show accuracy for lesion type, mass shape and margin,
and calcification morphology and distribution. Panel (f) reports the
percentage-point difference between the displayed results with and
without GRPO. These label-based scores evaluate lesion characterization
separately from the final assessment and free-form explanation.}
\label{fig:supp-descriptor-plots}
\end{figure}

\subsection{Qualitative Comparison of Retained Evidence}

\begin{figure*}[t]
\centering
\includegraphics[width=0.98\textwidth,height=0.79\textheight,keepaspectratio]{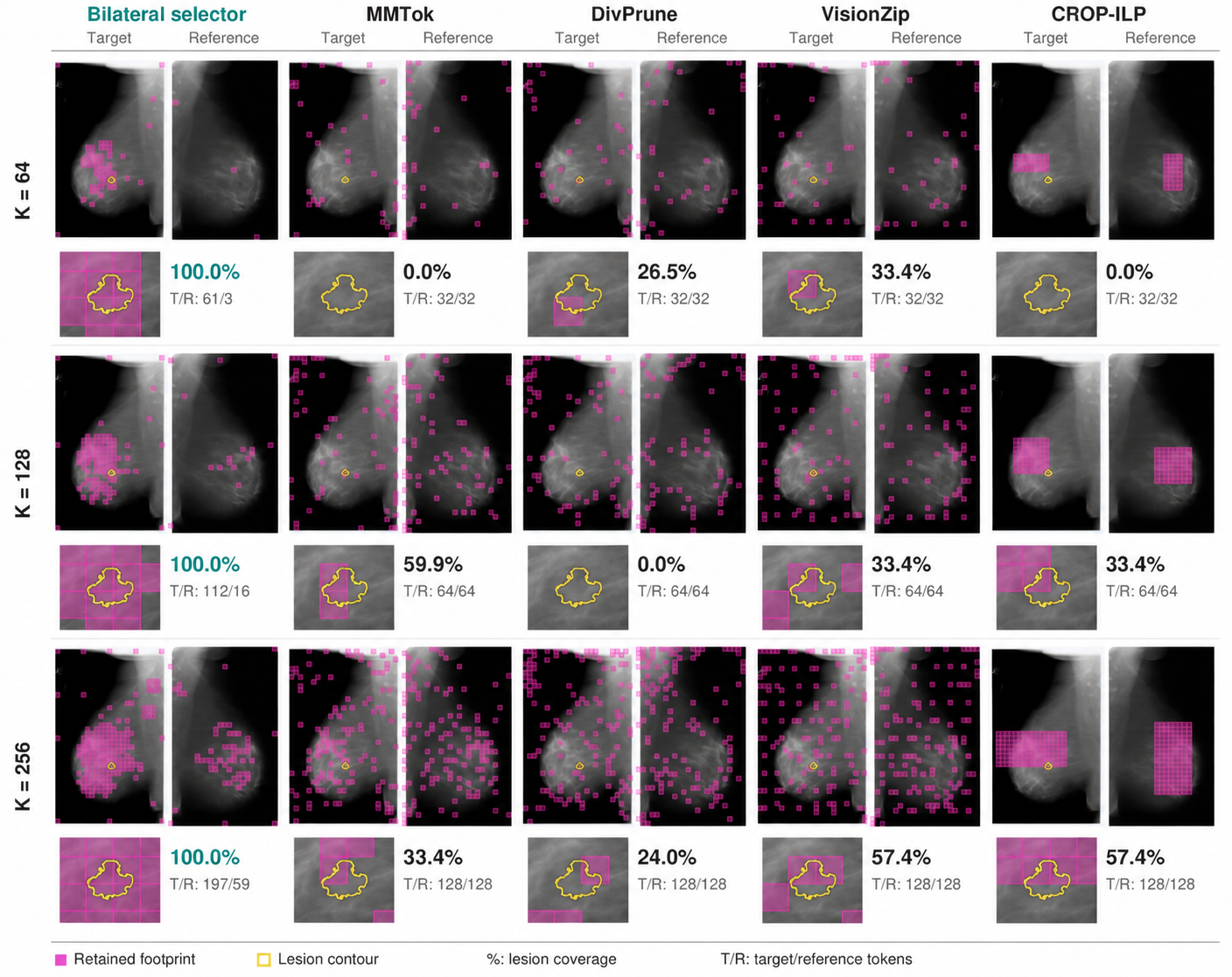}
\caption{\textbf{Visual-token selection across budgets for a benign-mass case.}
Rows correspond to $K=64,128,256$. Each method shows the target and
reference mammograms and a target-lesion close-up. Magenta marks
retained footprints; yellow outlines the annotated lesion. Percentages
denote lesion coverage, and T/R gives target/reference token counts.}
\label{fig:supp-token-selection}
\end{figure*}

Figure~\ref{fig:supp-token-selection} compares retained evidence for
one benign-mass example at $K\in\{64,128,256\}$. Both original
mammograms are shown for every method. The target close-up reveals
how each allocation covers the annotated lesion, while the T/R
counts show how the budget is divided between breasts.

In this example, the bilateral selector covers the entire annotated
lesion at all three budgets while allocating some tokens to the
reference breast. Other selections vary in their spatial coverage
as the budget changes.

\subsection{Lesion-Type and Attribute Prediction}

Figure~\ref{fig:supp-descriptor-plots} reports lesion-type and attribute
accuracy across the three budgets. The plotted GRPO comparison shows
positive changes in shape, margin, morphology, and distribution at each
budget, while lesion-type accuracy decreases at $K=64$ and $K=128$ and
increases at $K=256$. These findings suggest that improvements
in lesion characterization do not always coincide with better lesion-type classification.


\subsection{Predictions before and after GRPO}

\begin{figure*}[t]
\centering
\includegraphics[width=\textwidth,height=0.76\textheight,keepaspectratio]{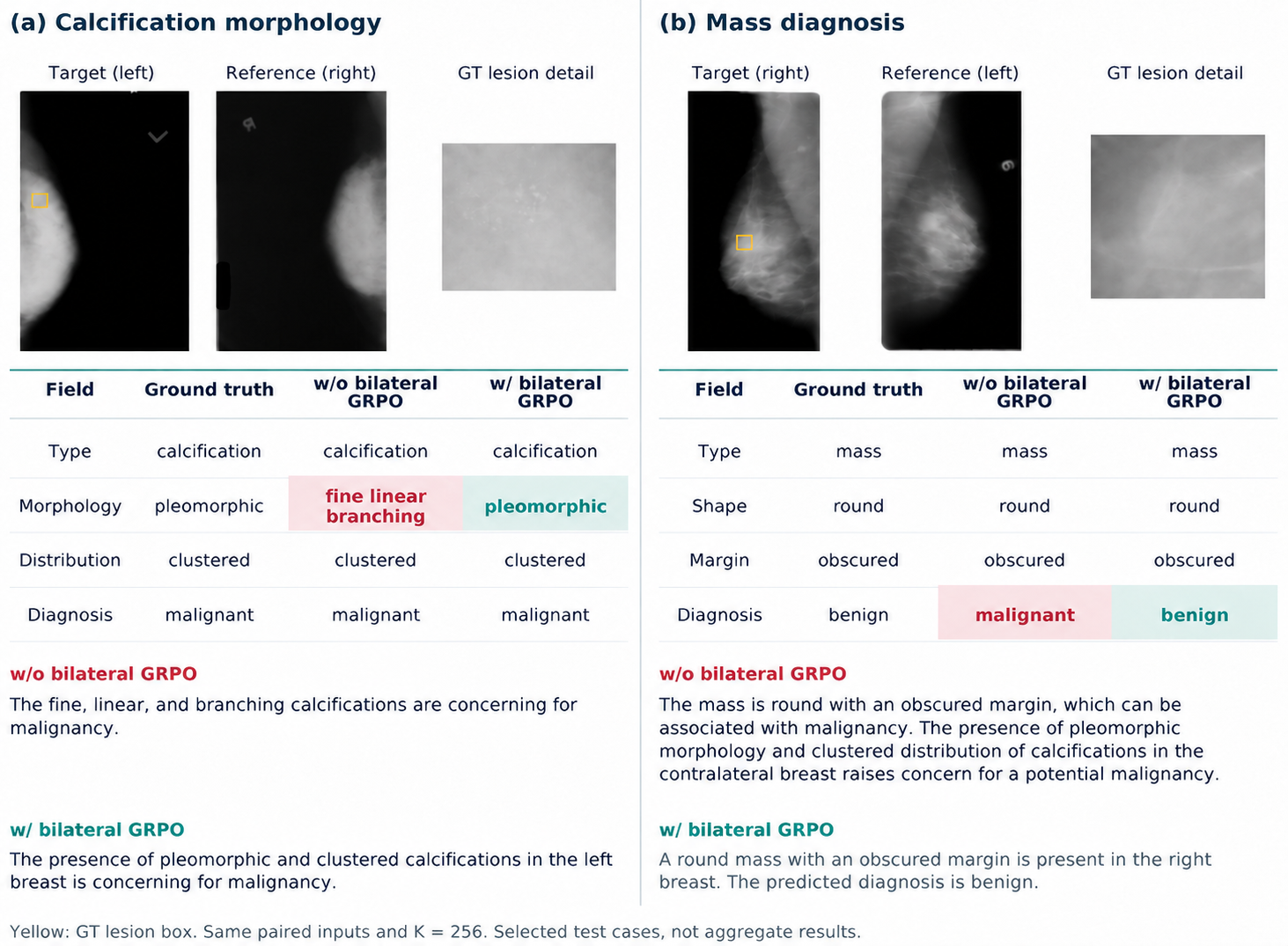}
\caption{\textbf{Selected predictions before and after Bilateral GRPO at $K=256$.}
In (a), the calcification morphology changes from fine linear branching
to the annotated pleomorphic category, while the malignant assessment
remains unchanged. In (b), the mass assessment changes from malignant
to benign, while shape and margin remain unchanged. Yellow boxes mark
the annotated lesions.}
\label{fig:supp-grpo-examples}
\end{figure*}

Figure~\ref{fig:supp-grpo-examples} shows two selected test cases from CBIS-DDSM at
$K=256$. The calcification example changes an attribute prediction
while preserving the correct assessment. The mass example changes the
assessment while preserving its predicted shape and margin. These
cases illustrate that descriptor and assessment correctness can change
separately. The lesion boxes and enlarged crops are for visualization.


\clearpage

\clearpage

{\small
\bibliographystyle{ieeenat_fullname}
\bibliography{references}

@inproceedings{alvar2025divprune,
  author = {Saeed Ranjbar Alvar and Gursimran Singh and Mohammad Akbari and Yong Zhang},
  title = {{DivPrune}: Diversity-based Visual Token Pruning for Large Multimodal Models},
  booktitle = {Proceedings of the IEEE/CVF Conference on Computer Vision and Pattern Recognition (CVPR)},
  pages = {9392--9401},
  year = {2025}
}

@article{bai2025qwen25vl,
  author = {Shuai Bai and Keqin Chen and Xuejing Liu and Jialin Wang and Wenbin Ge and Sibo Song and Kai Dang and Peng Wang and Shijie Wang and Jun Tang and Humen Zhong and Yuanzhi Zhu and Mingkun Yang and Zhaohai Li and Jianqiang Wan and Pengfei Wang and Wei Ding and Zheren Fu and Yiheng Xu and Jiabo Ye and Xi Zhang and Tianbao Xie and Zesen Cheng and Hang Zhang and Zhibo Yang and Haiyang Xu and Junyang Lin},
  title = {{Qwen2.5-VL} Technical Report},
  journal = {arXiv preprint arXiv:2502.13923},
  year = {2025}
}

@article{bannur2024maira2,
  author = {Shruthi Bannur and Kenza Bouzid and Daniel C. Castro and Anton Schwaighofer and Anja Thieme and Sam Bond-Taylor and Maximilian Ilse and Fernando P{\'e}rez-Garc{\'i}a and Valentina Salvatelli and Harshita Sharma and Felix Meissen and Mercy Ranjit and Shaury Srivastav and Julia Gong and Noel C. F. Codella and Fabian Falck and Ozan Oktay and Matthew P. Lungren and Maria Teodora Wetscherek and Javier Alvarez-Valle and Stephanie L. Hyland},
  title = {{MAIRA-2}: Grounded Radiology Report Generation},
  journal = {arXiv preprint arXiv:2406.04449},
  year = {2024}
}

@inproceedings{benartzi2025deepbirads,
  author = {Gil Ben-Artzi and Feras Daragma and Shahar Mahpod},
  title = {Deep {BI-RADS} Network for Improved Cancer Detection from Mammograms},
  booktitle = {Pattern Recognition: 27th International Conference, ICPR 2024},
  pages = {17--30},
  year = {2025}
}

@inproceedings{chen2024fastv,
  author = {Liang Chen and Haozhe Zhao and Tianyu Liu and Shuai Bai and Junyang Lin and Chang Zhou and Baobao Chang},
  title = {An Image is Worth 1/2 Tokens After Layer 2: Plug-and-Play Inference Acceleration for Large Vision-Language Models},
  booktitle = {European Conference on Computer Vision (ECCV)},
  year = {2024}
}

@inproceedings{chen2025mimo,
  author = {Yanyuan Chen and Dexuan Xu and Yu Huang and Songkun Zhan and Hanpin Wang and Dongxue Chen and Xueping Wang and Meikang Qiu and Hang Li},
  title = {{MIMO}: A Medical Vision Language Model with Visual Referring Multimodal Input and Pixel Grounding Multimodal Output},
  booktitle = {Proceedings of the IEEE/CVF Conference on Computer Vision and Pattern Recognition (CVPR)},
  pages = {24732--24741},
  year = {2025}
}

@inproceedings{dong2026mmtok,
  author = {Sixun Dong and Juhua Hu and Mian Zhang and Ming Yin and Yanjie Fu and Qi Qian},
  title = {{MMTok}: Multimodal Coverage Maximization for Efficient Inference of {VLMs}},
  booktitle = {International Conference on Learning Representations (ICLR)},
  year = {2026}
}

@inproceedings{guo2025crop,
  author = {Jiawei Guo and Feifei Zhai and Pu Jian and Qianrun Wei and Yu Zhou},
  title = {{CROP}: Contextual Region-Oriented Visual Token Pruning},
  booktitle = {Proceedings of the 2025 Conference on Empirical Methods in Natural Language Processing},
  pages = {9756--9772},
  year = {2025}
}

@InProceedings{Harmanani_2026_CVPR,
    author    = {Harmanani, Mohamed and Long, Bining and Guo, Zhuoxin and Wilson, Paul F.R. and Sabour, Amirhossein and To, Minh Nguyen Nhat and Fichtinger, Gabor and Abolmaesumi, Purang and Mousavi, Parvin},
    title     = {Vision-Language Models Encode Clinical Guidelines for Concept-Based Medical Reasoning},
    booktitle = {Proceedings of the IEEE/CVF Conference on Computer Vision and Pattern Recognition (CVPR) Findings},
    month     = {June},
    year      = {2026},
    pages     = {5589-5598}
}

@article{chen2025think,
  title={Think twice to see more: Iterative visual reasoning in medical vlms},
  author={Chen, Kaitao and Rui, Shaohao and Jiang, Yankai and Wu, Jiamin and Zheng, Qihao and Song, Chunfeng and Wang, Xiaosong and Zhou, Mu and Liu, Mianxin},
  journal={arXiv preprint arXiv:2510.10052},
  year={2025}
}

@inproceedings{huang2026medvlthinker,
  author = {Xiaoke Huang and Juncheng Wu and Hui Liu and Xianfeng Tang and Yuyin Zhou},
  title = {{MedVLThinker}: Simple Baselines for Multimodal Medical Reasoning},
  booktitle = {Proceedings of the Fifth Machine Learning for Health Symposium},
  publisher = {PMLR},
  pages = {384--398},
  year = {2026}
}

@incollection{heath1998current,
  title={Current status of the digital database for screening mammography},
  author={Heath, Michael and Bowyer, Kevin and Kopans, Daniel and Kegelmeyer Jr, P and Moore, Richard and Chang, Kyong and Munishkumaran, S},
  booktitle={Digital mammography: nijmegen, 1998},
  pages={457--460},
  year={1998},
  publisher={Springer}
}

@inproceedings{huang2026nuwa,
  author = {Yihong Huang and Fei Ma and Yihua Shao and Jingcai Guo and Zitong Yu and Laizhong Cui and Qi Tian},
  title = {{N\"uwa}: Mending the Spatial Integrity Torn by {VLM} Token Pruning},
  booktitle = {International Conference on Learning Representations (ICLR)},
  year = {2026}
}

@inproceedings{koh2020cbm,
  author = {Pang Wei Koh and Thao Nguyen and Yew Siang Tang and Stephen Mussmann and Emma Pierson and Been Kim and Percy Liang},
  title = {Concept Bottleneck Models},
  booktitle = {Proceedings of the 37th International Conference on Machine Learning},
  pages = {5338--5348},
  year = {2020}
}

@article{lai2026medr1,
  author = {Yuxiang Lai and Jike Zhong and Ming Li and Shitian Zhao and Yuheng Li and Konstantinos Psounis and Xiaofeng Yang},
  title = {{Med-R1}: Reinforcement Learning for Generalizable Medical Reasoning in Vision-Language Models},
  journal = {IEEE Transactions on Medical Imaging},
  volume = {45},
  number = {6},
  pages = {2727--2737},
  year = {2026}
}

@article{lee2017cbis,
  author = {Rebecca Sawyer Lee and Francisco Gimenez and Assaf Hoogi and Kanae Kawai Miyake and Mia Gorovoy and Daniel L. Rubin},
  title = {A curated mammography data set for use in computer-aided detection and diagnosis research},
  journal = {Scientific Data},
  volume = {4},
  pages = {170177},
  year = {2017}
}

@inproceedings{li2026medreasonr1,
  author = {Yifan Li and Fenghe Tang and Yingtai Li and Shaohua Kevin Zhou},
  title = {{MedReason-R1}: Learning to Reason for {CT} Diagnosis with Reinforcement Learning and Local Zoom},
  booktitle = {2026 IEEE 23rd International Symposium on Biomedical Imaging (ISBI)},
  pages = {1--5},
  year = {2026}
}

@inproceedings{liu2019cbn,
  author = {Yuhang Liu and Zhen Zhou and Shu Zhang and Ling Luo and Qianyi Zhang and Fandong Zhang and Xiuli Li and Yizhou Wang and Yizhou Yu},
  title = {From Unilateral to Bilateral Learning: Detecting Mammogram Masses with Contrasted Bilateral Network},
  booktitle = {Medical Image Computing and Computer Assisted Intervention -- MICCAI 2019},
  pages = {477--485},
  year = {2019}
}

@article{liu2022agn,
  author = {Yuhang Liu and Fandong Zhang and Chaoqi Chen and Siwen Wang and Yizhou Wang and Yizhou Yu},
  title = {Act Like a Radiologist: Towards Reliable Multi-View Correspondence Reasoning for Mammogram Mass Detection},
  journal = {IEEE Transactions on Pattern Analysis and Machine Intelligence},
  volume = {44},
  number = {10},
  pages = {5947--5961},
  year = {2022}
}

@article{liu2026breastgpt,
  author = {Yang Liu and Jiajin Zhang and Danyang Tu and Yaojun Hu and Jiao Qu and Jiuyu Zhang and Yu Shi and Wei Fang and Shi Gu and Ling Zhang and Yingda Xia},
  title = {{BreastGPT}: A Multimodal Large Language Model for the Full Spectrum of Breast Cancer Clinical Routine},
  journal = {arXiv preprint arXiv:2606.04911},
  year = {2026}
}

@inproceedings{luo2025vividmed,
  author = {Lingxiao Luo and Bingda Tang and Xuanzhong Chen and Rong Han and Ting Chen},
  title = {{VividMed}: Vision Language Model with Versatile Visual Grounding for Medicine},
  booktitle = {Proceedings of the 2025 Conference of the Nations of the Americas Chapter of the Association for Computational Linguistics: Human Language Technologies},
  pages = {1800--1821},
  year = {2025}
}

@article{nguyen2023vindr,
  author = {Hieu T. Nguyen and Ha Q. Nguyen and Hieu H. Pham and Khanh Lam and Linh T. Le and Minh Dao and Van Vu},
  title = {{VinDr-Mammo}: A large-scale benchmark dataset for computer-aided diagnosis in full-field digital mammography},
  journal = {Scientific Data},
  volume = {10},
  pages = {277},
  year = {2023}
}

@inproceedings{park2026medicad,
  author = {Woohyeon Park and Jaeik Kim and Sunghwan Steve Cho and Pa Hong and Wookyoung Jeong and Yoojin Nam and Namjoon Kim and Ginny Y. Wong and Ka Chun Cheung and Jaeyoung Do},
  title = {{Medic-AD}: Towards Medical Vision-Language Model's Clinical Intelligence},
  booktitle = {Proceedings of the IEEE/CVF Conference on Computer Vision and Pattern Recognition (CVPR)},
  pages = {36321--36331},
  year = {2026}
}

@inproceedings{shao2024visualcot,
  author = {Hao Shao and Shengju Qian and Han Xiao and Guanglu Song and Zhuofan Zong and Letian Wang and Yu Liu and Hongsheng Li},
  title = {Visual {CoT}: Advancing Multi-Modal Language Models with a Comprehensive Dataset and Benchmark for Chain-of-Thought Reasoning},
  booktitle = {Advances in Neural Information Processing Systems},
  year = {2024}
}

@article{shao2024deepseekmath,
  author = {Zhihong Shao and Peiyi Wang and Qihao Zhu and Runxin Xu and Junxiao Song and Xiao Bi and Haowei Zhang and Mingchuan Zhang and Y. K. Li and Y. Wu and Daya Guo},
  title = {{DeepSeekMath}: Pushing the Limits of Mathematical Reasoning in Open Language Models},
  journal = {arXiv preprint arXiv:2402.03300},
  year = {2024}
}

@inproceedings{wang2023disasymnet,
  author = {Xin Wang and Tao Tan and Yuan Gao and Luyi Han and Tianyu Zhang and Chunyao Lu and Regina Beets-Tan and Ruisheng Su and Ritse Mann},
  title = {{DisAsymNet}: Disentanglement of Asymmetrical Abnormality on Bilateral Mammograms using Self-adversarial Learning},
  booktitle = {Medical Image Computing and Computer Assisted Intervention -- MICCAI 2023},
  pages = {57--67},
  year = {2023}
}

@inproceedings{yang2025visionzip,
  author = {Senqiao Yang and Yukang Chen and Zhuotao Tian and Chengyao Wang and Jingyao Li and Bei Yu and Jiaya Jia},
  title = {{VisionZip}: Longer is Better but Not Necessary in Vision Language Models},
  booktitle = {Proceedings of the IEEE/CVF Conference on Computer Vision and Pattern Recognition (CVPR)},
  pages = {19792--19802},
  year = {2025}
}

@inproceedings{zhang2025sparsevlm,
  author = {Yuan Zhang and Chun-Kai Fan and Junpeng Ma and Wenzhao Zheng and Tao Huang and Kuan Cheng and Denis A. Gudovskiy and Tomoyuki Okuno and Yohei Nakata and Kurt Keutzer and Shanghang Zhang},
  title = {{SparseVLM}: Visual Token Sparsification for Efficient Vision-Language Model Inference},
  booktitle = {Proceedings of the 42nd International Conference on Machine Learning},
  pages = {74840--74857},
  year = {2025}
}

@article{zhu2026mammorg,
  author = {Jiayi Zhu and Fuxiang Huang and Yu Xie and Xi Wang and Zhixuan Chen and Yuan Guo and Qingcong Kong and Zhenhui Li and Qiong Luo and Hao Chen},
  title = {Cross-Modal Clinical Knowledge Integration for Mammography Report Generation},
  journal = {arXiv preprint arXiv:2605.31093},
  year = {2026}
}
}
\end{document}